\documentclass[superscriptaddress]{revtex4-2}
\usepackage{CJK}
\usepackage{amsmath,mathtools,amssymb}
\newcommand{\bd}[1]{\boldsymbol{#1}}
\let\revappendix\appendix

\begin{document}

\title{Learning normal form autoencoders for data-driven discovery of universal, parameter-dependent 
governing equations}
%nonlinear dynamics}% Force line breaks with \\
%\thanks{A footnote to the article title}%
\author{Manu~Kalia}
\email[Corresponding author: ]{m.kalia@utwente.nl}
\affiliation{Department of Applied Mathematics, University of Twente, P.O. Box 217, 7500 AE Enschede, The Netherlands}
\author{Steven L.~Brunton}%
% \email{sbrunton@uw.edu}
\affiliation{%
 Department of Mechanical Engineering,
 University of Washington,
 Seattle, WA 98195, USA}%
\author{Hil G.E.~Meijer}
% \email{h.g.e.meijer@utwente.nl}
\affiliation{Department of Applied Mathematics, University of Twente, P.O. Box 217, 7500 AE Enschede, The Netherlands}%
\author{Christoph~Brune}
% \email{c.brune@utwente.nl}
\affiliation{Department of Applied Mathematics, University of Twente, P.O. Box 217, 7500 AE Enschede, The Netherlands}%
\author{J. Nathan~Kutz}
%\email{kutz@uw.edu}
\affiliation{%
 Department of Applied Mathematics,
 University of Washington,
 Seattle, WA 98195, USA
}%
\date{\today}% It is always \today, today,
             %  but any date may be explicitly specified

\begin{abstract}
%%% PRL ABSTRACT
%We introduce deep learning autoencoders to discover coordinate transformations that capture the underlying parametric dependence of a dynamical system in terms of its canonical normal form, allowing for a simple representation of the parametric dependence and bifurcation structure. 
%The autoencoder constrains the latent variable to a given normal form, allowing it to learn the appropriate coordinate transformation. 
%We demonstrate the method on a number of examples, showing that it can capture normal forms associated with Hopf, pitchfork, transcritical and saddle node bifurcations.

%%% ARXIV ABSTRACT
Complex systems manifest a small number of instabilities and bifurcations that are canonical in nature, resulting in universal pattern forming characteristics as a function of some parametric dependence.
% %
Such parametric instabilities are mathematically characterized by their universal unfoldings, or normal form dynamics, whereby a parsimonious model can be used to represent the dynamics.
% %
Although center-manifold theory guarantees the existence of such low-dimensional normal forms, finding them has remained a long standing challenge.
% %
In this work, we introduce deep learning autoencoders to discover coordinate transformations that capture the underlying parametric dependence of a dynamical system in terms of its canonical \emph{normal form}, allowing for a simple representation of the parametric dependence and bifurcation structure. 
The autoencoder constrains the latent variable to adhere to a given normal form, thus allowing it to learn the appropriate coordinate transformation.
% %
We demonstrate the method on a number of example problems, showing that it can capture a diverse set of normal forms associated with Hopf, pitchfork, transcritical and/or saddle node bifurcations.
This method shows how normal forms can be leveraged as canonical and universal building blocks in deep learning approaches for model discovery and reduced-order modeling. 

\end{abstract}

\maketitle

\section{Introduction}
Instabilities and bifurcations in dynamical systems are canonical in nature, taking on a small but distinct number of forms that dominate pattern formation across every field of physics, engineering, and biology~\cite{cross1993pattern}. 
For such bifurcations, local equations exist that describe the {\em universal unfolding} of the change in qualitative behavior arising from parametric dependencies~\cite{guckenheimer2013nonlinear}. 
These equations, called \textit{normal forms}, are low-dimensional and depend on a minimal set of key parameters that modulate the dynamics.
Current methods for characterizing such instabilities require knowledge of the governing equations and asymptotic approximations in local neighborhoods of the state and parameter space~\cite{cross1993pattern,guckenheimer2013nonlinear}.
However, modern data-driven approaches aim to quantify global behavior directly from measurements, including capturing representations of normal forms~\cite{yair2017reconstruction}.
Physics-informed machine learning architectures~\cite{brunton2019data,raissi2019physics,Noe2019science,bar2019learning} leverage the flexibility and universal approximation capabilities of deep neural networks to learn characterizations of critical physics, including coordinate systems for the parsimonious representation of the dynamics~\cite{Brunton2016,champion2019data}.
However, deep learning approaches have typically focused on a single parameter regime, and they have not resulted in explicit parameterizations of bifurcations and instabilities in the dynamics.
In this work, we use deep learning to discover the low-dimensional coordinate system that encodes the underlying normal form dynamics and pattern-forming bifurcation structure of parameter-dependent high dimensional data, giving a data-driven, low-dimensional and universal representation of the dynamics.

Model discovery and model reduction methods aim to discover coordinate systems, or low-dimensional subspaces, in which high-dimensional data evolves.
Modal decomposition techniques, such as {\em proper orthogonal decomposition} (POD)~\cite{Benner2015} and {\em dynamic mode decomposition} (DMD)~\cite{Schmid2010}, approximate {\em linear} subspaces using dominant correlations in spatio-temporal data~\cite{Taira2007}.
Linear subspaces, however, are highly restrictive and ill-suited to handle parametric dependencies.
Attempts to circumvent these shortcomings include using multiple linear subspaces covering different temporal or spatial domains, 
diffusion maps~\cite{yair2017reconstruction,dietrich2020manifold,holiday2019manifold}, or more recently, using deep learning to compute underlying nonlinear subspaces which are advantageous for the representation of the dynamics~\cite{Brunton2016,champion2019data,linot2020deep,lee2020model}. 
Deep learning provides a flexible architecture for data representation, which has led to its significant integration into the physical and engineering sciences~\cite{brunton2019data,raissi2019physics}.
%
%The joint discovery of coordinates and dynamics has been used to learn simplified dynamical systems models, both linear and nonlinear~\cite{lusch2018deep,champion2019data,gin2019deep}.
%
Specifically, within such a framework, the {\em sparse identification of nonlinear dynamics} (SINDy) can uncover parsimonious nonlinear models~\cite{Brunton2016,Rudy2017}. Building on the SINDy framework, the goal here is to capture the underlying normal form that encodes the parametric dependence of the data and its underlying bifurcation.

\begin{figure*}[t]
	\centering
	\includegraphics[width=\textwidth]{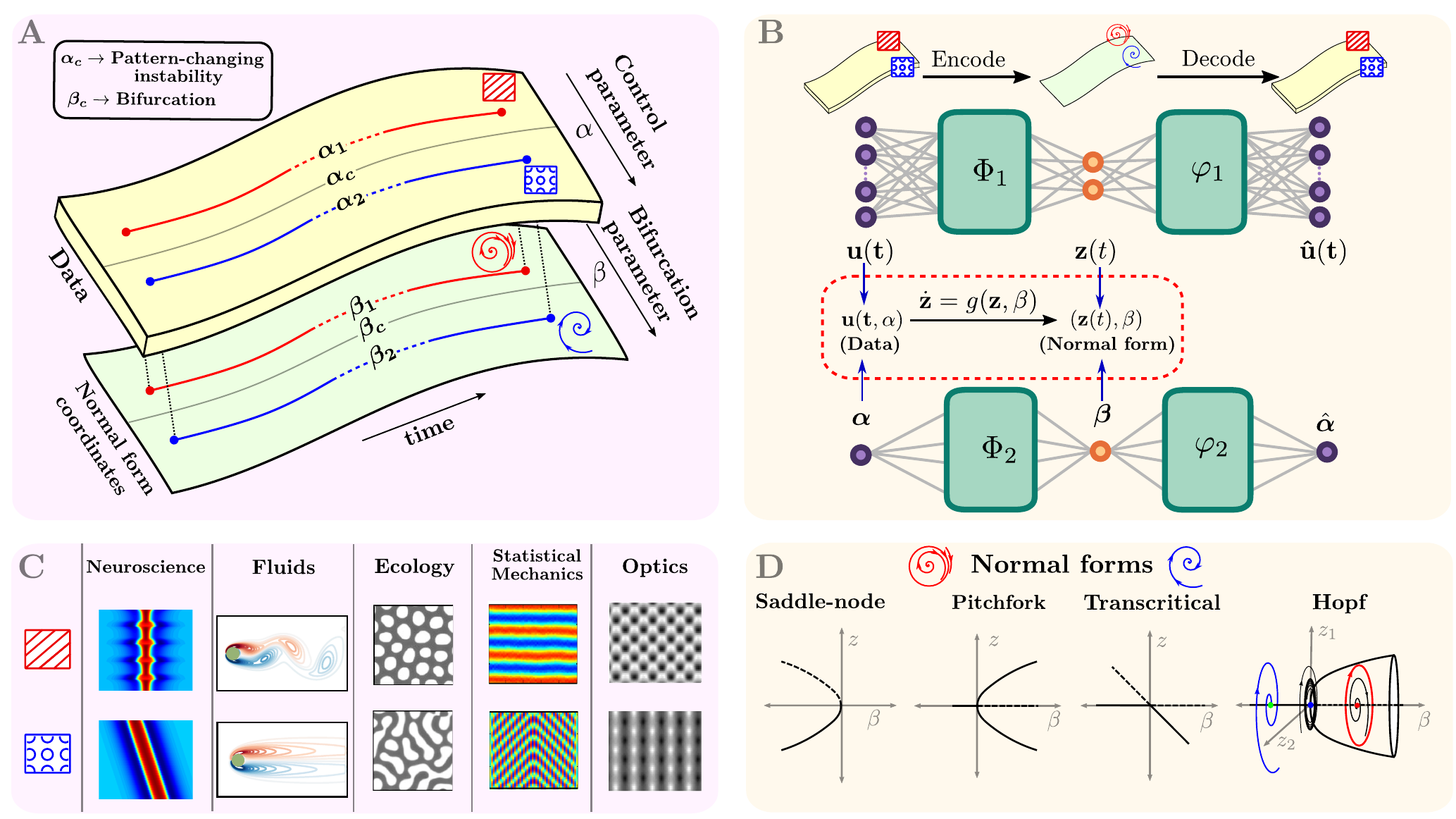}
	\caption{\label{fig:overview} Instabilities, or bifurcations, lead to pattern formation in various physical systems that are characterized by underlying normal forms. Parameterized data across an instability is considered, which arises from a physical system with control parameter $\alpha$ (top left). Such data is ubiquitous in the study of physical systems, for example, neuroscience~\cite{Coombes2005}, fluid physics, ecology~\cite{Fig1Ecology}, statistical physics~\cite{Fig1StatPhys} and optics~\cite{Fig1Optics} (bottom left). This data is collapsed down to the underlying normal form coordinates $(z,\beta)$, with bifurcation parameter $\beta$ using autoencoders (top right). The dynamics on the reduced coordinates $(z,\beta)$ are given by normal form equations. The different patterns from data are in one-to-one relation with the corresponding normal form patterns. Our novel approach uncovers a single parameterized equation (i.e., the normal form) that captures the parametric dependence across the data. This presents a plethora of normal forms to choose from, depending on the pattern changes observed in the data set (bottom right). Bottom left panel figures are reproduced with permission.}
\end{figure*}

Despite the diverse and rapid advancement of deep learning methods, the model discovery process has not yet captured the often simple parameter-dependence of the high-dimensional data, except with brute force parametrization. We highlight this issue in Fig.~\ref{fig:overview}-A. Consider a bifurcation occurring within data at a critical parameter $\alpha = \alpha_c$. This instability induces a dramatic change in the behavior of the system, yielding different patterns for parameter values before and after the bifurcation. Such changes are ubiquitous in physical systems (Fig.~\ref{fig:overview}-C) and present a challenge to cutting-edge model discovery methods. The different patterns are topologically inequivalent - they cannot be mapped onto each other by continuous, invertible transformations. Thus, observations from a single physical system yield irreconcilably different low-dimensional models, which is challenging for the aforementioned methods. Yet the underlying physics comes from a single model that simply walked through a bifurcation point. 

In this work, we present a deep learning approach that extracts low-dimensional coordinates from high-dimensional parameter-varying temporal data that exhibits instabilities. The coordinates and their parametric dependence are discovered using \textit{autoencoders} that transform observations of states and parameters simultaneously while constraining the transformed variables to the corresponding normal form equations, as shown in Fig.~\ref{fig:overview}-D. We demonstrate the method on various examples and instabilities: multiple bifurcations in a scalar ODE model, supercritical Hopf bifurcations in 1D partial differential equations (PDEs), and finally, a supercritical Hopf bifurcation in the 2D Navier-Stokes equations.

\subsection{Normal Forms and Bifurcations.}
The qualitative transitions in dynamics arising from bifurcations in temporal data are a cornerstone of dynamical systems analysis and \textit{bifurcation} theory~\cite{guckenheimer2013nonlinear,Kuznetsov2013}. 

Remarkably, there are only a small number of canonical instability types~\cite{cross1993pattern}, allowing us to understand a diversity of instabilities manifesting in nature. For instance, the Hopf normal form is $\dot{z} = (\beta + i\omega) z + z|z|^2,$ where the dot denotes time derivative. $\omega$ is the rotation frequency and $\beta$ is the bifurcation parameter which characterizes the crossing of a pair of complex conjugate eigenvalues moving from the left to right half plane in a linear stability analysis~\cite{cross1993pattern}. Thus, growth of an oscillatory field is expected. 

The Hopf bifurcation is only one of several bifurcations, with the simplest such bifurcations presented in Fig.~\ref{fig:overview}-D. These bifurcations describe the interactions between multiple steady states upon perturbing the parameter. For a scalar system, there are only three possible bifurcations. A stable equilibrium could collide with an unstable one before disappearing (limit point) or split into two stable equilibria with an unstable one in between (pitchfork). Lastly, colliding equilibria can be followed by reemergence of the stable-unstable pair of equilibria, but switched in position (transcritical). One of the most commonly observed bifurcations, is the Hopf bifurcation requiring a minimum of two state dimensions. Hopf, pitchfork, transcritical and saddle node bifurcations are the most commonly manifest instabilities of physical systems~\cite{cross1993pattern}.

Dynamical systems theory and center manifold theorems~\cite{guckenheimer2013nonlinear} provide conditions for the existence of low-dimensional subspaces, or center manifolds, where the dynamics of the projection of the original dynamics, is given by the normal form. These theorems guarantee that a high-dimensional system $\dot{\bd u}=f(\boldsymbol{u},\alpha), \boldsymbol{u} \in \mathbb{R}^n, \ n \gg 1$ depending on a parameter $\alpha$ exhibit generically a low dimensional model, typically one or two dimensional.
Moreover, these theorems state that if $\bd u(\alpha)$ exhibits a saddle node bifurcation, there exists a smooth invertible transformation $\varphi$ such that the dynamics of $\varphi(\boldsymbol{u}(\alpha))$ are given by the saddle node normal form. Thus center manifold theorems give guarantees that low-dimensional coordinates can be constructed using an appropriate normal form transformation $\varphi$. 

\subsection{Deep Learning of Normal Forms.} 
Consider a high-dimensional system $u(x,t;\alpha)$ parameterized by $\alpha$ where $x$ denotes space, $t$ denotes time. Discretizing $u$ along $n$ spatial locations $x$ gives the vector $\boldsymbol{u}(t) \in \mathbb{R}^n$. Further discretizing along $d$ timepoints gives the dataset $\boldsymbol{U} \in \mathbb{R}^{n \times d}$ composed of columns $\mathbf{u}(t)$ for $t = t_1,\dots,t_d$. Note that the data set $\mathbf{U}$ is parameterized by the parameter $\alpha$. The data measures a local instability at $\alpha=\alpha_c \in \mathbb{R}$. The objective is to extract low dimensional coordinates $\boldsymbol{z} \in \mathbb{R}^m, \ m \ll n$ and $\beta \in \mathbb{R}$ such that the dynamics of $\boldsymbol{z}$ are given by the normal form of the instability, 
\begin{equation}
    \dot{\bd z} = g(\boldsymbol{z}(t),\beta).
    \label{eq:nf}
\end{equation}
The coordinates $\boldsymbol{z}$ are extracted by constructing smooth, invertible transformations $\varphi_1$ and $\varphi_2$ such that 
\begin{equation}
   \boldsymbol{z}(t) = \varphi_1\boldsymbol{u}(t) \textnormal{ and } \beta = \varphi_2\alpha, \ \forall t.
   \label{eq:latent}
\end{equation}
We compute the functions $\varphi_1$ and $\varphi_2$ using deep learning. In particular, $\varphi_1$ and $\varphi_2$ are represented as fully connected neural networks. Further, we simultaneously compute neural networks $\psi_1$ and $\psi_2$ such that
\begin{equation}
    \psi_1\varphi_1(\boldsymbol{u}) \approx \boldsymbol{u} \textnormal{ and } \psi_2\varphi_2(\alpha) \approx \alpha
    \label{eq:autoencoder}
\end{equation}
to make $\varphi_j$ invertible. Such an approach is now standard in deep learning theory, and the pair $(\varphi_j,\psi_j)$ is collectively referred to as an \textit{autoencoder}~\cite{Hinton1994}.

Figure~\ref{fig:overview}-B shows the two autoencoders $(\varphi_j,\psi_j), \ j=1,2$, corresponding to the state and parameter respectively. Combining equations \eqref{eq:nf} and \eqref{eq:latent} gives,  
\begin{equation} 
    \dot{\bd z} = \mathrm{d}/\mathrm{dt} (\varphi_1\boldsymbol{u}) = (\nabla_{\bd u}\varphi_1)\dot{\bd u} = g(\varphi_1\boldsymbol{u}, \varphi_2\alpha). 
\end{equation}
This relation is exploited to constrain the two autoencoders to the normal form \eqref{eq:nf}. 
This is accomplished by computing minimizers of a loss function $\mathcal{L}$ that takes in the high-dimensional dataset $\boldsymbol{u}$, the neural networks $\varphi_j, \psi_j$ and the parameter $\alpha$,
\begin{equation} 
    \hat{\varphi}_j,\hat{\psi}_j = \arg \min_{\Theta} \mathcal{L}(\boldsymbol{u},\alpha,\varphi_j,\psi_j) = \arg \min_{\Theta} \sum_k \mathcal{L}_k,
\label{eq:consistency}
\end{equation}
where $\Theta$ is the large set of parameters underlying the autoencoders $(\varphi_j,\psi_j)$. 
The various terms $\mathcal{L}_k$ are outlined as follows. Terms $\mathcal{L}_{1,2}$ are the {\em autoencoder loss} terms that ensure $\varphi_j$ and $\psi_j$ are inverses of each other, enforcing Eq.~(\ref{eq:autoencoder}):
\begin{equation}
    \mathcal{L}_1 = \lambda_1 \|\boldsymbol{u} - \psi_1\varphi_1\boldsymbol{u}\|_2^2, \,\, 
    \mathcal{L}_2 = \lambda_2 \|\alpha - \psi_2 \varphi_2\alpha\|_2^2.
    %\stepcounter{equation}\tag{\theequation} 
    \nonumber
\end{equation}
The {\em consistency loss} terms $\mathcal{L}_{3,4}$ constrain the autoencoders to the condition Eq.~(\ref{eq:consistency}) and are given by,
\begin{align*}
    \mathcal{L}_3 &= \lambda_3\|\nabla_{\bd u}\dot{\bd u} - g(\varphi_1\boldsymbol{u},\varphi_2\alpha)\|_2^2, \\
    \mathcal{L}_4 &= \lambda_4 \|\dot{\bd u} - (\nabla_{\bd z} \psi_1)g(\varphi_1\bd u, \varphi_2\alpha)\|_2^2.
    %\stepcounter{equation}\tag{\theequation}
    \nonumber
\end{align*}    
Lastly, the {\em orientation loss} terms $\mathcal{L}_{5,6}$ ensure proper affine translation of the coordinates $(\boldsymbol{u},\beta)$ with respect to the original coordinates $(\boldsymbol{u},\alpha)$ and are given by,
\begin{align*}
    \mathcal{L}_5 = \lambda_5\|\mathbb{E}_t\varphi_1\boldsymbol{u}\|_2^2, \,\,
    \mathcal{L}_6 = \lambda_6 \|\textnormal{sgn}(\alpha) \pm \textnormal{sgn}(\varphi_2 \alpha)\|_2^2,
    %\stepcounter{equation}\tag{\theequation}
    \nonumber
\end{align*}
where $\mathbb{E}_t$ denotes expectation over the entire time trace. The neural networks require training data in order to learn the autoencoder structure. The training data consists of dynamical trajectories where the initial conditions and parameters are chosen from a uniform distribution. They are shuffled and paired together and then used together to simulate trajectories. Once trajectories are computed, they are divided into training and testing datasets. The testing dataset is used to assess the performance of the autoencoder scheme, while training data is used to learn the neural networks. The neural networks $\varphi_1$ and $\psi_1$ require $\bd u$ and $\dot{\bd u}$ as input, while $\varphi_2$ and $\psi_2$ take $\alpha$ as input. They are then trained using the ADAM optimizer~\cite{ADAM2017} for a fixed choice of parameters $\lambda_i$. For each of the examples presented ahead, details on training and validation, choice of neural networks, and regularization parameters $\lambda_i$ can be found in the Appendix. 

\begin{figure}[t]
\centering
\includegraphics[width=0.45\textwidth]{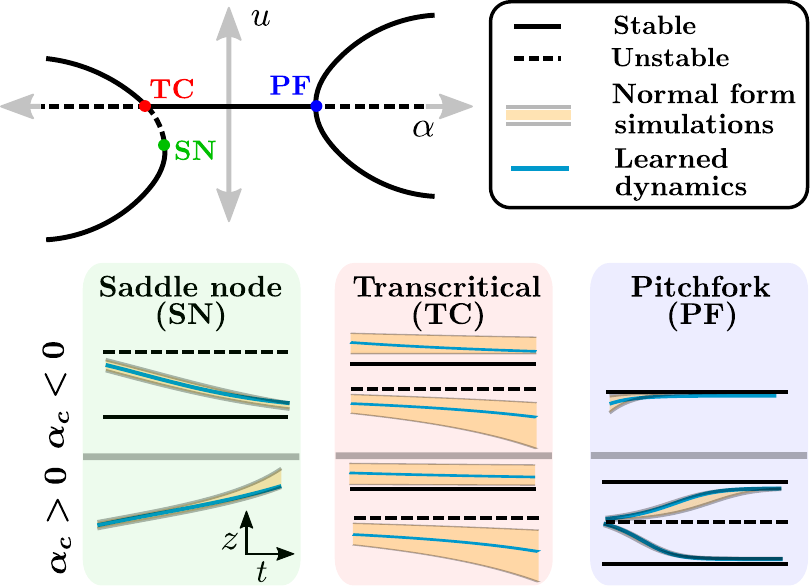}
\caption{\label{fig:results_scalar} Learned normal form coordinates for the various bifurcations present in the 1D system Eq.~\ref{eq:scalar}. For each bifurcation, traces from learned test samples are plotted (in blue) against an ensemble of simulations of the underlying normal form (yellow).}
\end{figure}

\section{Results}
In this section we demonstrate our method on four nonlinear dynamical systems: a scalar ODE, a neural field equation, the Lorenz96 equations and the Navier Stokes equation solved on a 2D spatial domain.

\subsection{Scalar ODE system}
The autoencoder scheme is first demonstrated on a system that exhibits the three scalar normal forms introduced in Figure~\ref{fig:overview}-C. The system is characterized by a scalar ODE, given by 
\begin{equation}
    \dot{u} = \gamma u(\alpha-\alpha_{\it pf}-u^2)(\alpha-\alpha_{\it sn}+(u-u_{\it sn})^2), u \in \mathbb{R}, 
        \label{eq:scalar}
\end{equation}
where $\gamma = 0.01$, $u_{\it sn} = \alpha_{\it sn} = -6$ and $\alpha_{\it pf} = 6$. The bifurcation diagram of Eq.~\eqref{eq:scalar} in Fig.~\ref{fig:results_scalar} shows how all the different scalar bifurcations are distributed in parameter space. Our objective is to use data generated from Eq.~\ref{eq:scalar} and constrain it to each of the three individual normal forms. For each bifurcation scenario, data is collected from the neighborhood of a bifurcation and then constrained to the respective normal form using the autoencoder. A total of 500 initial conditions $(u,\alpha)$ are sampled per bifurcation scenario and used for training. Results are presented in Fig.~\ref{fig:results_scalar}. For each bifurcation, samples from test data are transformed to $(z,\beta)$ coordinates using $\varphi_{1,2}$ and plotted against time (blue) for different $\alpha$. An ensemble of simulations (in yellow) of the normal form are used for comparison. The learned coordinates $(z,\beta)$ show remarkable agreement with the normal form, for each of the three bifurcation scenarios. 
%[\textbf{Put relative norms here}].

\begin{figure}[t]
\centering
\includegraphics[width=0.8\textwidth]{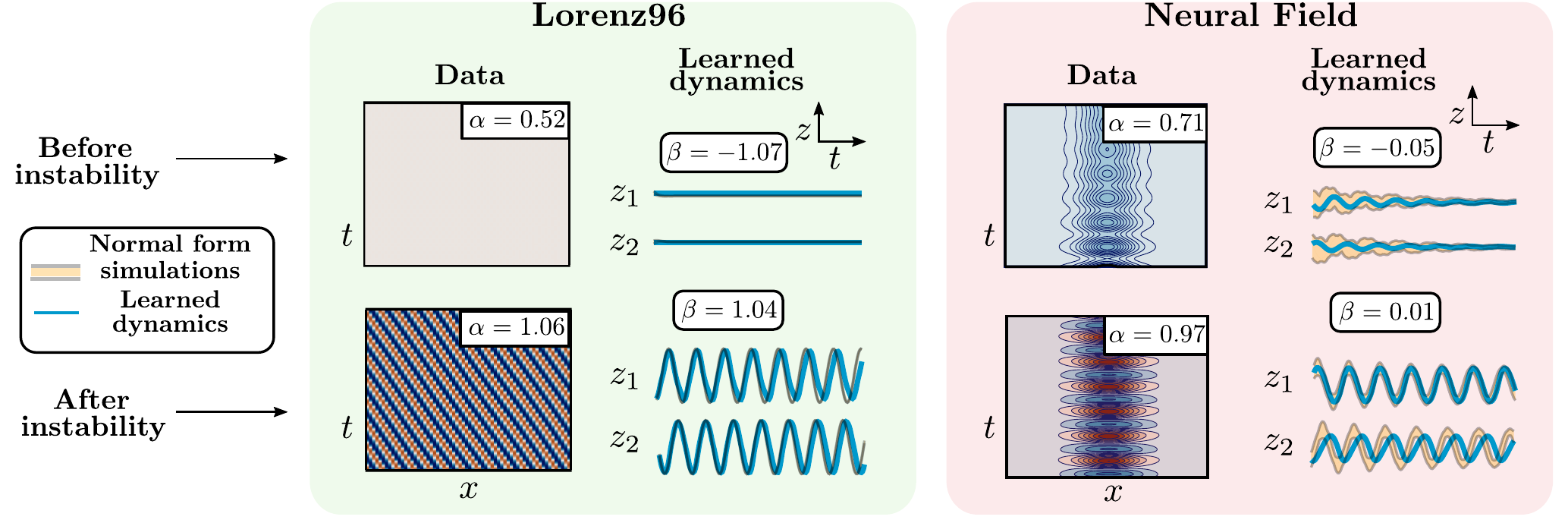}
\caption{\label{fig:results_high} Learned Hopf normal form coordinates for the two high-dimensional systems, Lorenz96 (Eq.\ref{eq:Lorenz96}) and Neural Field (Eq.~\ref{eq:neuralfield}), using test dataset samples. In both cases, imulations $(\boldsymbol{u},\alpha)$ are shown alongside the learned normal form coordinates $(\boldsymbol{z},\beta)$ (in blue) for values of $\alpha$ on both sides of the Hopf bifurcation point. For comparison, the Hopf normal form is simulated and plotted in the background (yellow).}
\end{figure}

Next, we consider two 1D spatio-temporal systems, that exhibit supercritical Hopf bifurcations, the Lorenz96 equations~\cite{lorenz2006} and the neural field equations~\cite{Amari1977}. Although the bifurcation is the same, the pattern formation is different. In the Lorenz96 case, the Hopf bifurcation manifests in a travelling wave pattern~\cite{VanKekem2018}, while an oscillatory bump solution, called a `breather', emerges in the neural field equation~\cite{Folias2011}, see Fig.~\ref{fig:results_high}.

\subsection{Lorenz96 system.} 
The Lorenz96 equations~\cite{lorenz2006} are widely used in model discovery and data assimilation. The equations are given by
\begin{equation}
    \dot{u}_j = -u_{j-1}(u_{j-2}-u_{j+1}) - u_j + \alpha, \ u \in \mathbb{R}^n
        \label{eq:Lorenz96}
\end{equation}
for $j = 1,2,3...n$ with boundary conditions $u_1 = u_{n}$ and $u_2 = u_{n-1}$. For $n=64$, the trivial equilibrium $\boldsymbol{u} = \alpha$ undergoes a supercritical Hopf bifurcation with respect to $\alpha$ at $\alpha = \alpha_0 = 0.84975$~\cite{VanKekem2018}. Across the bifurcation, a stationary solution transits to a moving stripe pattern, which is interpreted as a travelling wave solution. We sampled $10^3$ initial conditions $(\boldsymbol{u},\alpha)$ to train the neural networks. Results using test data are shown in Fig.~\ref{fig:results_high}. The learned coordinates $\boldsymbol{z}$ (in blue) are two-dimensional and match well with the simulated Hopf normal form (yellow) on both sides of the bifurcation.

\subsection{Neural field equation.} The neural field equations describe the neuronal potential for a one-dimensional continuum of neural tissue~\cite{Amari1977,Wilson1972}. The dynamics due to an input inhomogeneity lead to a Hopf bifurcation of a stationary pattern leading to breathers when varying the input strength~\cite{Folias2011, Coombes2005}. The governing equations are
\begin{equation} 
	\dot{u}  = -u - \kappa a + (w * f(u)) + I(x), \, \dot{a}  =  (u-a) /\tau_{\it nf}.
\label{eq:neuralfield}
\end{equation}
The operator $*$ represents a spatial convolution with $w(x) \equiv w(x-y) = w_e\exp(-((x-y)/\sigma_e)^2)$ the spatial connectivity kernel and $f(u)$ is a sigmoid given by $f(u) = \left( 1+ \exp( \beta_{\it nf}(u-u_{\it thr}) ) \right)$ transforming the potential $u$ into a firing rate. The spatially non-uniform input $I(x)$ is given by $I(x) = \alpha \exp(-(x/\sigma)^2)$ where $\kappa \!=\! 2.75, \tau_{\it nf} \!=\! 10, w_e \!=\! 1, \sigma_e \!=\! 1, \beta_{\it nf} \!=\! 6, u_{\it thr} \!=\! 0.375, \sigma \!=\! 1.2$.

A supercritical Hopf bifurcation with respect to a stationary bump response occurs at $\alpha=0.8040$. In contrast to the Lorenz96 case, the stationary bump solution transits to an asymptotically stable periodic solution~\cite{Coombes2005}. States $(u,a)$ are discretized over a uniform spatial grid of size $64$ each. Then, $10^3$ initial conditions $(\boldsymbol{u},\alpha)$ are used to generate training data. The normal form autoencoder results are presented in Fig.~\ref{fig:results_high}. Even though the pattern formation in this example is different from Lorenz96, the learned coordinates match the Hopf normal form behavior again.  
\begin{figure*}[t]
\centering
\includegraphics[width=\textwidth]{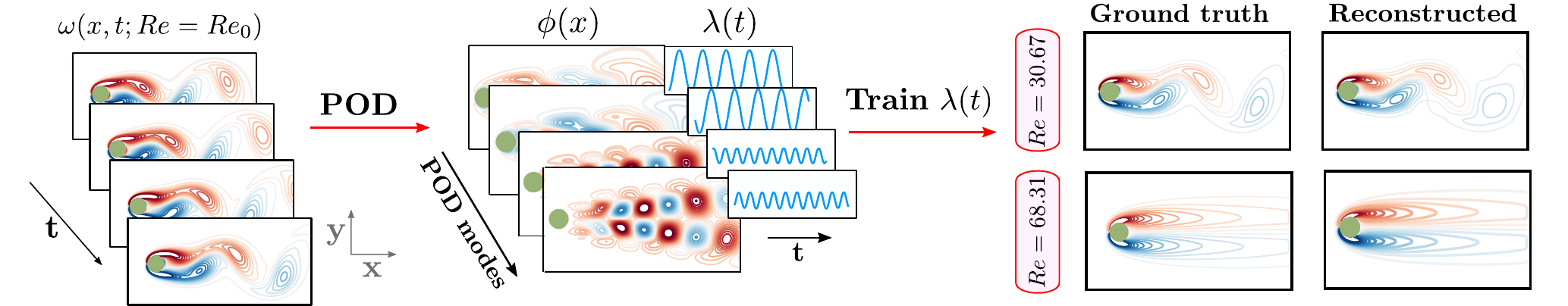}
\caption{\label{fig:results_fluid} Training the normal form autoencoder over POD data generated from fluid flow past a cylinder. First, Eq.~\ref{eq:navierstokes} is solved using 250 initial conditions $(\omega,Re)$ generated across the vortex shedding instability~\cite{Bearman69}. Then proper orthogonal decomposition is performed to obtain spatial modes $\phi(x)$ and their temporal coefficients $\lambda(t)$, which forms a low-dimensional dynamical system, which is used to train the normal form autoencoder.}
\end{figure*}

The 1D PDEs considered are relatively low dimensional systems. For such systems, training fully connected neural network-based autoencoders is feasible. However, considering higher-dimensional PDEs leads to the so-called `curse of dimensionality', and using fully connected neural networks is no longer feasible. In this situation, one has two options: use neural networks designed to assuage the curse of dimensionality, or reduce the dimension of data prior to training. In the next example, we choose the latter.

\subsection{Fluid flow in 2D.} As a more challenging example, we simulate the fluid flow past a circular cylinder with the two-dimensional, incompressible Navier-Stokes equations: 
\begin{equation}
%    \begin{aligned}
\nabla \cdot \boldsymbol{u} =0, \,\,\,\, 
\partial_{t} \boldsymbol{u}+(\bd{u} \cdot \nabla)\bd{u}=-\nabla p+\frac{1}{R e} \Delta \boldsymbol{u}
\label{eq:navierstokes}
%\end{aligned}
\end{equation}
where $\boldsymbol{u}$ is the two-component flow velocity field in 2D and $\boldsymbol{p}$ is the pressure term.  
For Reynold's number $Re = Re_c \approx 47$, the fluid flow past a cylinder undergoes a supercritical Hopf bifurcation, where the steady flow for $Re<Re_c$ transitions to unsteady vortex shedding~\cite{Bearman69}. The unfolding of the transition gives the celebrated Stuart-Landau ODE, which is essentially the supercritical Hopf normal form written in complex coordinates, and this has resulted in accurate and efficient reduced-order models for this system~\cite{Noack2003jfm,Noack2011book}. 

The scalar vorticity field $\omega \equiv \nabla \times \boldsymbol{u}$ is useful in reducing the complexity of the problem to a single component per grid point. The Hopf bifurcation persists in the vorticity field, and we use the vorticity field to construct datasets.
Datasets are generated over a 2D spatial grid of $487 \times 250$ points across the domain $[-2,10] \times [-3,3]$. The discretization results in $121750$ grid points, illustrating the aforementioned curse of dimensionality. This is mitigated by restricting the dataset to a lower dimension using proper orthogonal decomposition~\cite{brunton2019data}. 

The normal form autoencoder provides a fundamentally different approach to characterizing the low-dimensional dynamics than Galerkin projection of the governing equations onto POD modes~\cite{Noack2003jfm,Noack2011book,Benner2015}. First, Eq.~\ref{eq:navierstokes} is solved and the vorticity field $\omega(t,x)$ is computed. The reduced-order dynamical system $\lambda(t)$ is derived from the method of snapshots. Here, $\lambda \in \mathbb{R}^4$. This dynamical system characterizes the temporal evolution of spatial modes $\phi(x)$, that are kept aside. An example of spatial modes $\phi(x)$ and their temporal coefficients $\lambda(t)$ are presented in Fig.~\ref{fig:results_fluid}. The temporal coefficients $\lambda(t)$ are then used to train the neural networks. The training dataset is generated from 250 initial conditions $(\omega,Re)$. The learned coordinates $(\boldsymbol{z},\beta)$ show agreement with the Hopf normal form dynamics (not shown). Further, the spatial modes $\phi(x)$ are used to project the learned coordinates $(\boldsymbol{z},\beta)$ back to the vorticity field $\omega$ using the relation $\hat{\omega} \approx \bar{\omega} +  \sum_j\phi_j(x)(\psi_1\varphi_1\lambda_j(t))$, where $\bar{\omega}$ is the time-averaged solution. The reconstructed vorticity field $\hat{\omega}$ shows remarkable agreement with ground truth for cases on either side of the bifurcation, demonstrating good agreement of the learned $\psi_1$ with the inverse of the encoder $\varphi_1$. 

\section{Conclusion}
We demonstrated how to use deep learning to discover a coordinate transformation in which dynamics can be directly characterized in terms of universal normal form descriptions. Such embeddings of parameter-dependent dynamics automate many of the theoretical constructions used to characterize spatio-temporal pattern forming systems~\cite{cross1993pattern}. Indeed, the architecture leverages the vast body of knowledge concerning the small number of canonical instabilities that emerge in diverse models of physics, biology, and engineering. Our approach is currently limited by the vast state-parameter space required to be sampled to learn the whole phase space properly. Moreover, the approach currently requires a priori knowledge of the observed bifurcation, and this can perhaps be remedied with an offline detection step in the dataset. Model discovery techniques like SINDy~\cite{Brunton2016} can then be leveraged to identify a bifurcation based on a library of normal forms.  

Our approach has consequences for dynamical systems theory and data-driven model discovery alike. The approach can be extended to discover underlying low-dimensional, reduced-order models and center-manifold reductions using normal forms as the fundamental building blocks. Moreover, the method makes use of theoretical guarantees that allow such embeddings to exist. The flexibility of the autoencoder provides a modeling framework for finding the required coordinate transformations to the low-rank, universal unfolding of the dynamics.

All of the code used to produce the results presented in this work is available publicly on GitHub at: \\ \verb|github.com/dynamicslab/NormalFormAE|. 

\section{Acknowledgements}
MK was  supported by the Deutsche Forschungsgemeinschaft, FOR2795 `Synapses under stress' (Ro2327/14-1) and the NDNS travel grant NDNST2020001. SLB acknowledges funding support from the US Army Research Office (ARO W911NF-19-1-0045; Program manager, Dr. Matthew Munson. CB acknowledges funding support from the EU Horizon 2020 research and innovation programme under the Marie-Sklodowska-Curie grant agreement No 777 826. JNK acknowledges funding support from the Air Force Office of Scientific Research (FA9550-19-1-0011). All authors declare no competing interests.
\newpage

\revappendix*
\section{} 
%\documentclass[aps,prl,superscriptaddress]{revtex4-2}
%\usepackage{a4wide}
%\usepackage{mathtools,amsmath,amssymb}
%\usepackage[ruled,vlined]{algorithm2e}
%\newcommand{\bd}[1]{\boldsymbol{#1}}
%\title{Supplemental Material - Learning normal form autoencoders for data-driven discovery of universal, parameter-dependent governing equations}
%\begin{document}
%\section*{Supplementary Material}
This appendix describes the data acquisition, preprocessing and neural network training for the normal form autoencoder approach.

\subsection*{Data acquisition for training} For each of the examples, datasets $\boldsymbol{U},\dot{\boldsymbol{U}}$ and $\boldsymbol{\alpha}$ are constructed, which serve as inputs to the neural networks. These datasets are obtained by solving the corresponding governing equation. In this work, the governing equation is characterized by a parameterized, smooth, autonomous differential equation
\begin{equation}
    \dot{u}= f(u,x; \alpha), \ u \in \mathbb{R}^n, n\geq 1.
\end{equation}
The solution $\boldsymbol{u} \equiv [u(x_1,t),\dots,u(x_N,t)]^T$ is defined on a finite spatiotemporal grid $(x,t)$ using the initial value $(\boldsymbol{u}_{0}, \alpha_0)$. The datasets $\boldsymbol{U},\dot{\boldsymbol{U}}$ and $\boldsymbol{\alpha}$ are constructed by concatenating several solutions $\boldsymbol{u}$ for a collection of initial values $(\boldsymbol{u}_{0}, \alpha_0)$. First, we define datasets 
\begin{align*}
    \bd{U}^{(j)} &\equiv \bd{U}(\alpha = \alpha^{(j)}) = \begin{bmatrix}
    | &  &| \\
\boldsymbol{u}_{0}^{(j)} & \dots &\boldsymbol{u}_{t_f}^{(j)} \\
    | &  &|
    \end{bmatrix}  , \\
        \dot{\bd U}^{(j)} &\equiv \dot{\bd U}(\alpha = \alpha^{(j)}) = \begin{bmatrix}
    | &  &| \\
    \bd{f}(\bd{u}_0^{(j)}; \alpha^{(j)}) & \dots & \bd{f}(\bd{u}_{t_f}^{(j)}; \alpha^{(j)}) \\
    | &  &|
    \end{bmatrix},
    \tag{\theequation}\stepcounter{equation}
\end{align*}
where $t_f$ denotes the final time point. The function $\bd f(\bd u; \alpha)$ is defined by vectorizing $f$ over the discrete spatial grid $x$,
\begin{equation}
    \bd f(\bd u; \alpha) = \begin{bmatrix}
        | \\ f(\bd{u},x_j) \\ |
    \end{bmatrix}.
\end{equation}
The solution $\bd{U}^{(j)}$ may contain transients to the steady state solution, which are removed. Trimming off transients allows sampling of trajectories closer to steady state solutions, resulting in better conformity to normal form dynamics.

\noindent Next, $N$ initial conditions $(\bd u^{(j)}_0,\alpha^{(j)})$ are stacked together to get the datasets $\bd U, \dot{\bd U}$ and $\bd \alpha$,
\begin{equation}
    \bd{U} = 
    \begin{bmatrix}
        \bd{U}^{(0)} & \dots & \bd{U}^{(N)} 
    \end{bmatrix} \textnormal{, }
    \dot{\bd U} = 
    \begin{bmatrix}
        \dot{\bd U}^{(0)} & \dots & \dot{\bd U}^{(N)}
    \end{bmatrix} \textnormal{ and }
    \bd{\alpha} = 
    \begin{bmatrix}
        \alpha_0 & \dots & \alpha_N
    \end{bmatrix}
\end{equation}
Datasets 
\begin{align*}
    \bd{X}_{\it train} &= \{\bd{U}^{(\it train)},\dot{\bd U}^{(\it train)},\alpha^{(train)}\},\\
    \bd{X}_{\it test} &= \{\bd{U}^{(\it test)},\dot{\bd U}^{(\it test)},\alpha^{(test)}\},
\end{align*}
are constructed for training and validation, respectively.  The results presented in the main manuscript are performed over validation sets for each example.

\subsubsection*{Choosing initial values $(\bd u_0,\alpha_0)$}
The initial condition $(\bd{u}_0,\alpha_0)$ is sampled from a uniform distribution $\mathcal{U}$ based on the domain $[-1,1]$ such that points from either side of the bifurcation parameter $\alpha_c$ are uniformly sampled. First, parameters $\sigma_u$ and $\sigma_\alpha$ are fixed, such that
\begin{align*}
    \bd{u}_0 &= \bd{u}_c + \sigma_u \mathcal{U}[-1,1],\\
    \alpha_0 &= \alpha_c + \sigma_\alpha\mathcal{U}[-1,1]
    \stepcounter{equation}\tag{\theequation}\label{eq:datagen}
\end{align*}
where $\bd{u} = \bd{u}_c$ is the steady state (equilibrium) at which the bifurcation occurs.

\subsection*{Training}
The neural networks $\varphi_{1,2},\psi_{1,2}$ are fully connected neural networks with a single activation function active in the hidden layers only. In this work, we use the hyperbolic tangent ($\mathrm{tanh}$, for all Hopf examples) and exponential linear unit ($\mathrm{elu}$, for scalar ODE examples) functions as activation, as they allow for the transformed data to be smoothly equivalent to the original dataset. This choice makes the corresponding encoder and decoder smooth. After training, $\psi \circ \varphi \approx I$, which makes $\varphi$ approximately a diffeomorphism. Using center manifold theory \cite{Kuznetsov2013}, we thus obtain the existence of feasible solutions to the neural network problem.

\vspace{3mm}
The input to the normal form autoencoder is the set $\bd{X}_{\it train}$, as introduced earlier. In the latent space, we get
\begin{align*}
    \bd{Z} &= \varphi_1\bd{U}, \\
    \bd{\beta} &= \varphi_2\bd{\alpha},
    \tag{\theequation}\stepcounter{equation}
\end{align*}
where we drop the subscript $(\it train)$ for notational convenience. Passing the two latent variables through the decoder gives
\begin{align*}
    \bd{\hat{U}} &= \psi_1\bd{z}, \\
    \bd{\hat{\alpha}} &= \psi_2\bd{\beta}.
    \tag{\theequation}\stepcounter{equation}
\end{align*}
We also compute $\dot{\hat{\bd U}}$ and $\dot{\bd{z}}$ in order to compute the consistency loss terms. This is done via the chain rule as follows
\begin{align*}
    \dot{\bd z} &= \frac{\mathrm d}{\mathrm dx}(\varphi_1\bd{U}) \\
             &= (\nabla_{\bd{u}}\varphi_1)\dot{\bd U}. 
\end{align*}
In order to compute $\dot{\hat{\bd U}}$, we first compute the time-derivative estimate of the latent variable, $\dot{\hat{\bd z}}$ which is given by 
$$
\dot{\hat{\bd z}} = g(\bd{z},\bd{\beta}).
$$
This gives $\dot{\hat{\bd U}}$ via the relation
\begin{equation}
    \dot{\hat{\bd U}} = (\nabla_{\bd{z}}\psi_1)\dot{\hat{\bd z}}.
\end{equation}
The loss function $\mathcal{L}$ is thus given by,
\begin{equation}
    \mathcal{L} = \sum_j\mathcal{L}_j
\end{equation}
where,
\begin{align*}
    \mathcal{L}_{1} &= \lambda_1 \frac{1}{Nt_f}\sum_k \|\bd{u}_k - \bd{\hat{u}}_k\|_2^2 = \lambda_1 \frac{1}{Nt_f} \sum_k \|\bd{u}_k - \psi_1\varphi_1\bd{u}_k\|_2^2, \\
    \mathcal{L}_{2} &= \lambda_2 \frac{1}{Nt_f}\sum_k \|\bd{\alpha}_k - \bd{\hat{\alpha}}_k\|_2^2 = \lambda_2 \frac{1}{Nt_f} \sum_k \|\bd{\alpha}_k - \psi_2\varphi_2\bd{\alpha}_k\|_2^2, \\
    \mathcal{L}_3 &= \lambda_3 \frac{1}{Nt_f}\sum_k\|\dot{\hat{\bd z}}_k - \dot{\bd z}_k\|_2^2 = \lambda_3 \frac{1}{Nt_f}\sum_k \|(\nabla_{\bd u}\varphi_1)\dot{u}_k - g(\varphi_1\bd u_k,\varphi_2 \bd \alpha_k)\|_2^2,\\
    \mathcal{L}_4 &= \lambda_4 \frac{1}{Nt_f}\sum_k\|\dot{\hat{\bd u}}_k - \dot{\bd u}_k\|_2^2 = \lambda_4 \frac{1}{Nt_f}\sum_k \|\dot{u}_k - (\nabla_{\bd u}\varphi_2)g(\varphi_1\bd u_k,\varphi_2 \bd \alpha_k)\|_2^2,\\
    \mathcal{L}_5 &= \lambda_5 \frac{1}{N}\left\|\frac{1}{t_f}\sum_k \bd u_k\right\|_1 = \lambda_5\frac{1}{N}\|\mathbb{E}_t\bd U\|_1, \\
    \mathcal{L}_6 &= \lambda_6 \frac{1}{Nt_f}\left\|\textrm{sgn}\alpha - \textrm{sgn}\beta\right\|_1 = \lambda_6\frac{1}{Nt_f}\| \textrm{sgn}\alpha - \textrm{sgn}(\varphi_1\alpha) \|_1.
\end{align*}

Once the loss function is computed, the set of neural network parameters $\Theta$, comprising both autoencoders, are simultaneously trained using the ADAM optimizer \cite{kingma2017adam} with learning rate $\eta$, whose value depends on the example. For all examples, the \verb|Flux.jl| package in the Julia language is used to train the neural networks. All code is available online at \verb|github.com/dynamicslab/NormalFormAE|. For visualization, the latent dynamics $\bd z$ are plotted for all samples in the validation dataset. Using a uniform distribution of initial conditions centered around $(\bd z_0,\beta) = (\varphi_1\bd u_0,\alpha)$, an ensemble of simulations of the normal form are generated and plotted in the background. The neural networks are trained repeatedly over the batches of data generated (called epochs) till the fit to the simulations in the latent space stabilizes.

\subsubsection*{Orientation loss terms $\mathcal{L}_{5,6}$}
The loss terms $\mathcal{L}_{5,6}$ ensure that the latent variables $(\bd z,\beta)$ are properly oriented with respect to the normal form, and are hence called \textit{orientation loss} terms. Generically, the state variables in normal forms are scaled such that the bifurcation occurs at $(\bd z,\beta) = (0,0)$. Loss term $\mathcal{L}_5$ ensures that the time average of the latent space of a simulation is constrained to 0. This is pertinent specifically to the supercritical Hopf normal form, where the stable equilibrium for $\beta<0$ is at $z=0$, and the stable periodic orbit for $\beta>0$ is centered around the now unstable equilibrium $z=0$. The loss term $\mathcal{L}_6$ ensures that the direction of the bifurcation in the latent space is consistent with that of the normal form.

\subsubsection*{Choice of regularization constants $\lambda_i$}
The choice of regularization parameters $\lambda_j$ depends on the example in consideration. They remain fixed for the entire training procedure. However, parameters $\lambda_3$ and $\lambda_4$ are the most sensitive and generally require testing by training the architecture for short epochs before making a final choice. The parameter $\lambda_3$ controls the fit of the latent space to the normal form, while the parameter $\lambda_4$ makes sure that the reconstructed data $\hat{\bd u}$ fits up to the first-order time derivative. For large values of these two parameters, the training procedure prioritizes the latent space fit, which in practice results in the latent variables converging to the solution $\bd z=0$. On the other hand, for very small values of $\lambda_3$ and $\lambda_4$, the latent space does not match well with the normal form simulations. As a rule of thumb, we choose $\lambda_{3,4}$ such that the corresponding loss terms $\mathcal{L}_{3,4}$ are a factor $10^{-2}$ of the autoencoder loss term $\mathcal{L}_1$. This choice prioritizes the term $\mathcal{L}_1$ slightly more, as done in \cite{Champion2019}, which is beneficial as the autoencoder fit for $(\varphi_1,\psi_1)$ is typically the slowest moving loss term during training iterations. For large values of $\lambda_5$, the solution $\bd z=0, \dot{\bd z}=0$ is prioritized. In order to avoid this, $\lambda_{5}$ is kept low. 

\subsubsection*{Scaling time with $\tau$}
This work deals with projecting dynamics $\bd u$ onto a low dimensional manifold such that the dynamics $\bd u$ on such a manifold obey a specific normal form equation. However, this introduces a time scale problem when dealing with finite time trajectories $\bd u$. Let us assume that data $\bd U$ corresponds to a Hopf bifurcation. For $\alpha>0$ close to $\alpha_c$, the period of the resulting periodic orbit is $T_{\alpha} \approx 2\pi/\omega_u$, where $\omega_u$ is the imaginary part of the center eigenvalue of the linearization of the dynamical system $\dot{u}=f$, at the $\alpha_c$. Any diffeomorphism of such a signal will preserve the period if the periodic orbit persists. Thus the corresponding latent variable $\bd Z$ would also have period $T_{\alpha}$. However, the period $T_{\beta}$ corresponding to the Hopf normal form for $\beta=\varphi_2\alpha$ would be different, as $\omega_z \neq \omega_u$. Thus, we introduce a time scaling parameter $\tau$ to mitigate the difference in the period. This is done by introducing a new time $t^*$ such that,
\begin{equation}
t^* = \tau^2 t,
\end{equation}
which gives a scaled normal form equation
\begin{equation}
    \frac{\mathrm{d}}{\mathrm dt^*}z = \frac{1}{\tau^2}g(z,\beta).
\end{equation}
The time scaling parameter $\tau$ is included in the neural network parameter set $\Theta$ and learnt simultaneously. However, for the supercritical Hopf bifurcation examples, $\tau$ can be approximated theoretically and thus does not need to be trained. The value of $\tau$ is set to
\begin{equation}
    \tau = \sqrt{T_\alpha/T_\beta},
\end{equation}
where $T_\alpha$ and $T_\beta$ are estimates of the period approximated from data $\bd U$ and Hopf normal form simulations, respectively, for parameters close to the bifurcation value. These estimates are readily made using Fourier transforms of the simulated time traces after removing transients to the periodic steady state.

\begin{table}[hbtp]
\begin{tabular}{|l|l|}
\hline             
\multicolumn{2}{|c|}{\textbf{Data generation}}\\
\hline             
Normal Form & Hopf \\
$u_c$ & $u=\alpha$ \\
$\alpha_c$ & $0.84975$ \\
$\sigma_u$ & $0.1$ \\
$\sigma_\alpha$ & $0.5$ \\
Time domain & $[0,80]$ \\
Space domain & $[-32,32]$, 64 points \\
$t_{\it size}$ & 500 \\
Training set size & 1000 \\
Test set size & 20 \\
Trim & First 200 points \\
\hline             
\end{tabular}      
 \begin{tabular}{|l|l|}
     \hline        
 \multicolumn{2}{|c|}{\textbf{Training parameters}} \\
 \hline            
$\lambda_1$ & $1$  \\
$\lambda_2$ & $10^{-2}$ \\
$\lambda_3$ & $10^{-3}$ \\
$\lambda_4$ & $10^{-3}$ \\
$\lambda_5$ & $0$ \\
$\lambda_6$ & $10^{-1}$ \\
 Batchsize &  $100$ \\
 $\bd U_{\it batch}$ dimension & $64 \times 30000 $ \\
 $\bd Z_{\it batch}$ dimension & $2 \times 30000$ \\
 $\varphi_1$ hidden layers & [32,16] \\
 $\psi_1$ hidden layers & [16,32] \\
 $\varphi_2$ hidden layers & [16,16] \\
 $\psi_2$ hidden layers & [16,16] \\
 $\tau$ & 0.825 \\
 $\eta_{\it ADAM}$ & $10^{-4}$ \\
 Epochs & 1000 \\
\hline
\end{tabular}
\caption{Data generation and training parameters for the Lorenz96 example.}
\end{table}

\begin{table}[hbtp]
\begin{tabular}{|l|l|}
\hline
\multicolumn{2}{|c|}{\textbf{Data generation}}\\
\hline
Normal Form & Hopf \\
$u_c$ & Not analytical \\
$\alpha_c$ & $0.8040$ \\
$\sigma_u$ & $0.1$ \\
$\sigma_\alpha$ & $0.5$ \\
Time domain & $[0,100]$ \\
Space domain & $[-6,6]$, 64 points \\
$t_{\it size}$ & 250 \\
Training set size & 1000 \\
Test set size & 20 \\
Trim & First 50 points \\
\hline
\end{tabular}
 \begin{tabular}{|l|l|}
     \hline
 \multicolumn{2}{|c|}{\textbf{Training parameters}} \\
 \hline
$\lambda_1$ & $1$  \\
$\lambda_2$ & $10^{-2}$ \\
$\lambda_3$ & $10^{-4}$ \\
$\lambda_4$ & $0$ \\
$\lambda_5$ & $10^{-3}$ \\
$\lambda_6$ & $0$ \\
 Batchsize &  $250$ \\
 $\bd U_{\it batch}$ dimension & $128 \times 50000 $ \\
 $\bd Z_{\it batch}$ dimension & $2 \times 50000$ \\
 $\varphi_1$ hidden layers & [64,32] \\
 $\psi_1$ hidden layers & [32,64] \\
 $\varphi_2$ hidden layers & [16,16] \\
 $\psi_2$ hidden layers & [16,16] \\
 $\tau$ & 1.4\\
 $\eta_{\it ADAM}$ & $10^{-4}$ \\
 Epochs & 2000 \\
\hline
\end{tabular}
\caption{Data generation and training parameters for the Neural Field example. As the critical equilibrium point is not analytical, this is computed by allowing the simulation to stabilize after $t\gg1$.}
\end{table}

\begin{table}[hbtp]
\begin{tabular}{|l|l|}
\hline
\multicolumn{2}{|c|}{\textbf{Data generation}}\\
\hline
Normal Form & Hopf \\
$u_c$ & Not analytical \\
$Re_c$ & $44.6$ \\
$\sigma_u$ & $10^{-2}$ \\
$Re$ domain & $[30,70]$,240 points \\
Time domain & $[0,77]$ \\
Space domain  & $[-2,10] \times [-3,3]$, $487 \times 250$ points \\
$t_{\it size}$ & 6180 \\
Training set size & 220 \\
Test set size & 20 \\
Trim & First 3250 points \\
\hline
\end{tabular}
 \begin{tabular}{|l|l|}
     \hline
 \multicolumn{2}{|c|}{\textbf{Training parameters}} \\
 \hline
$\lambda_1$ & $1$  \\
$\lambda_2$ & $1$ \\
$\lambda_3$ & $10^{-4}$ \\
$\lambda_4$ & $10^{-4}$ \\
$\lambda_5$ & $0$ \\
$\lambda_6$ & $10^{-1}$ \\
 Batchsize &  $110$ \\
 $\bd U_{\it batch}$ dimension & $4 \times 32230 $ \\
 $\bd Z_{\it batch}$ dimension & $2 \times 32230$ \\
 $\varphi_1$ hidden layers & [20,20,30] \\
 $\psi_1$ hidden layers & [20,20,20] \\
 $\varphi_2$ hidden layers & [10,10] \\
 $\psi_2$ hidden layers & [10,10] \\
 $\tau$ & 0.6\\
 $\eta_{\it ADAM}$ & $10^{-3}$ \\
 Epochs & 2700 \\
\hline
\end{tabular}
\caption{Data generation and training parameters for the Navier Stokes example.}
\end{table}

\subsection*{Demonstrated systems}
This section elaborates on the systems that were used to demonstrate the normal form autoencoder approach in the main manuscript and provides explicit training and validation details.

\subsubsection*{1D model}
\begin{figure}
    \includegraphics[width=0.6\textwidth]{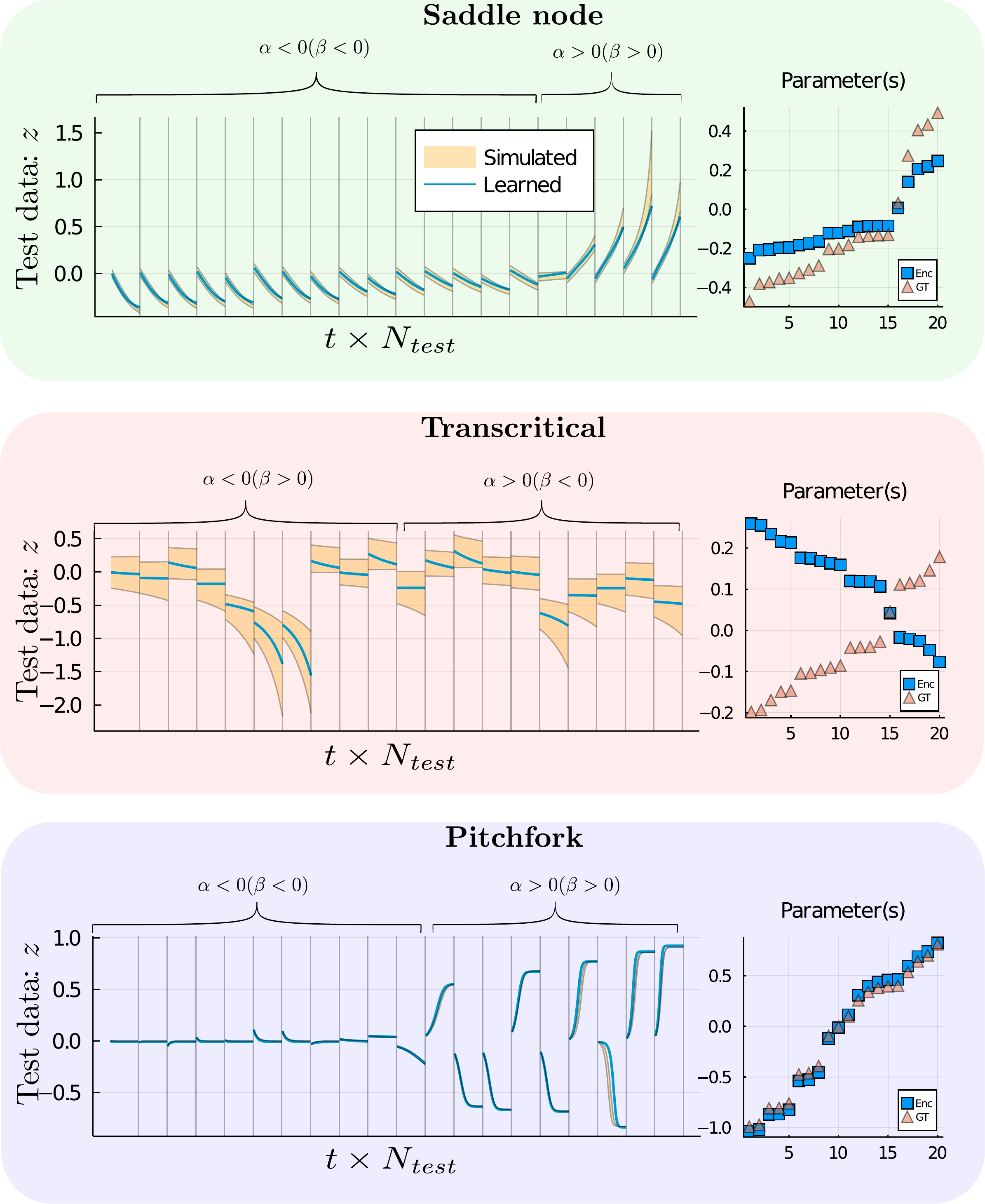}
    \caption{Validation results post-training for the scalar ODE example. The learned normal form coordinates computed via the formula $z = \varphi_1(\bd U)$ is plotted against time $t$ (in blue) for several simulations $N_{\it test}=20$ in the validation set (test data). The different simulations are separated from each other by a vertical gray line. The simulation of the respective normal form is plotted in the background in yellow, which represents an ensemble of 20 trajectories with initial values chosen from a uniform distribution around the first time point $\varphi_1\bd u_0$ and parameter $\beta=\varphi_2\alpha$. For each example the corresponding ground truth parameters ($\alpha$, in blue) and learned parameters ($\beta$, in orange) are also shown. Note that the parameter signs for the transcritical bifurcation are flipped as the direction of the bifurcation is in the reverse direction with respect to the normal form. For the transcritical case, the orientation term $\lambda_6$ is kept 0. }
    \label{fig:suppscalar}
\end{figure}

The scalar ODE explored is a toy model constructed to include all the major scalar bifurcations: saddle-node, pitchfork and transcritical. Data $\bd X_{\it train}$ is collected from the vicinity of each of the bifurcation points and used to train the normal form autoencoder separately for each bifurcation scenario. The system is given by 
\begin{equation}
    \dot{u} = \gamma u(\alpha-\alpha_{\it pf}-u^2)(\alpha-\alpha_{\it sn}+(u-u_{\it sn})^2), u \in \mathbb{R}, 
        \label{eq:suppscalar}
\end{equation}
where $\gamma = 0.01$, $x_{\it sn} = \alpha_{\it sn} = -6$ and $\alpha_{\it pf} = 6$. The three bifurcations occur at:
\begin{itemize}
\item Saddle node: $(u,\alpha) = (u_{\it sn},\alpha_{\it sn})$ 
\item Pitchfork: $(u,\alpha) = (0,\alpha_{\it pf})$
\item Transcritical: $(u,\alpha) = (0,\alpha_{\it sn} - u_{\it sn}^2)$
\end{itemize}
Before constructing the dataset $\bd X$, the system~(\ref{eq:suppscalar}) is first translated such that the bifurcation in consideration occurs at $(u,\alpha) = (0,0)$. For example, in the case of the pitchfork bifurcation, the translation $(u,\alpha) \mapsto (u,\alpha+\alpha_{\it pf})$ results in the new system,
\begin{equation}
    \dot{u} = \gamma u (\alpha - u^2)(\alpha+\alpha_{\it pf}-\alpha_{\it sn} + (u-u_{\it sn})^2).
    \label{eq:translated}
\end{equation}
The third term in the above equation is positive for sufficiently small $|\alpha|$. Thus, equation~(\ref{eq:translated}) is smoothly equivalent to the pitchfork normal form $\dot{u} = u(\alpha-u^2)$ \cite{Kuznetsov2013}. Smooth equivalence preserves orbits and the direction of time, but not the speed. We observe that the pitchfork normal form is scaled by a positive function $h(u,\alpha)$ given by,
\begin{equation}
    h(u,\alpha)= \gamma(\alpha+\alpha_{\it pf}-\alpha_{\it sn} + (u-u_{\it sn})^2).
\end{equation}
Thus, 
\begin{align*}
    \dot{u} &= \gamma u (\alpha - u^2)(\alpha+\alpha_{\it pf}-\alpha_{\it sn} + (u-u_{\it sn})^2) \\
            &= h(u,\alpha) u (\alpha-u^2) \\ 
            &\approx \frac{1}{\tau^2}u(\alpha-u^2), \ \tau \neq 0 \textnormal{  for } |u|,|\alpha| \textnormal{ sufficiently small}.
\end{align*}
This parameter $\tau$ is precisely the time scaling parameter introduced before, which we learn simultaneously with the neural network parameters while training. It can be shown for all other bifurcation scenarios that such a scaling would be necessary, and thus we learn the parameter $\tau$ for each case individually. Training results are shown in Fig.~\ref{fig:suppscalar}

\subsubsection*{Lorenz96}
The Lorenz96 system \cite{VanKekem2018,lorenz2006} is given by,
\begin{equation}
    \dot{u}_j = -u_{j-1}(u_{j-2}-u_{j+1}) - u_j + \alpha, \ u \in \mathbb{R}^n,
        \label{eq:suppL96}
\end{equation}
for $j=1,2 \dots n$ with boundary conditions $u_1 = u_{n}$ and $u_2 = u_{n-1}$. In this work, $n=64$, for which a supercritical Hopf bifurcation occurs at the trivial equilibrium $\bd u = \alpha$, for $\alpha=0.84975$. As done in the previous section, the system is translated such that the bifurcation occurs at the origin. Moreover, for all choice of parameter $\alpha$, it is made sure that the equilibrium occurs at the origin $\bd u=0$. This is done via the translation $(\bd u,\alpha) \mapsto (\bd u + \alpha,\alpha + \alpha_c)$, where $\alpha_c$ is the bifurcation point $\alpha=0.84975$.

\vspace{3mm}
The system is solved with $1000$ initial conditions $(\bd u,\alpha)$ over a temporal domain $[0,80]$, with $500$ time points per simulation. The transients to the travelling wave pattern state are removed by neglecting the first $200$ points of the simulation. The training set $\bd X_{\it train}$ thus comprises of $1000$ simulations, giving rise to $3\times 10^5$ training samples. For each training iteration, a batch of $100$ simulations is used, giving rise to 10 training iterations per epoch. The system is trained for $1000$ epochs or $10^4$ training iterations. 

\subsubsection*{Neural Field}
The neural field equations \cite{Amari1977,Wilson1972} are a system of integrodifferental equations that describe the dynamics of electrical activity in spatially continuous neural tissue. In this work, we consider a specific formulation of neural field equations with an input inhomogeneity that manifests in a Hopf bifurcation of a stationary pattern leading to breathers when varying the input strength~\cite{Folias2011, Coombes2005}. The governing equations are
\begin{equation} 
	\dot{u}  = -u - \kappa a + (w * f(u)) + I(x), \quad \dot{a} = (u-a) /\tau_{\it nf}.
\label{eq:suppnf}
\end{equation}
Here again, we translate the system such that the equilibrium and the bifurcation point both occur at the origin. However, in contrast to the Lorenz96 example, an analytical expression for the bifurcating equilibrium is absent. This is approximated from data. First, the parameter is translated to the bifurcation point $\alpha \mapsto \alpha+\alpha_c$, where $\alpha_c = 0.804$ is the bifurcation point. The system is then solved for several initial conditions $(\bd u,\alpha)$. Next, for $\alpha<0$, the last point $\bd u_{t_f}$ of the simulation to be the equilibrium $\bd u_{\it eq}$ and for $\alpha>0$, the time average of the simulation $\mathbb{E}_t\bd{U}^{(j)}$, after ignoring transients, is chosen as the equilibrium $\bd u_{\it eq}$. Then the translation $\bd u \mapsto \bd u + \bd u_{\it eq}$ is performed.

\begin{figure}
    \includegraphics[width=0.6\textwidth]{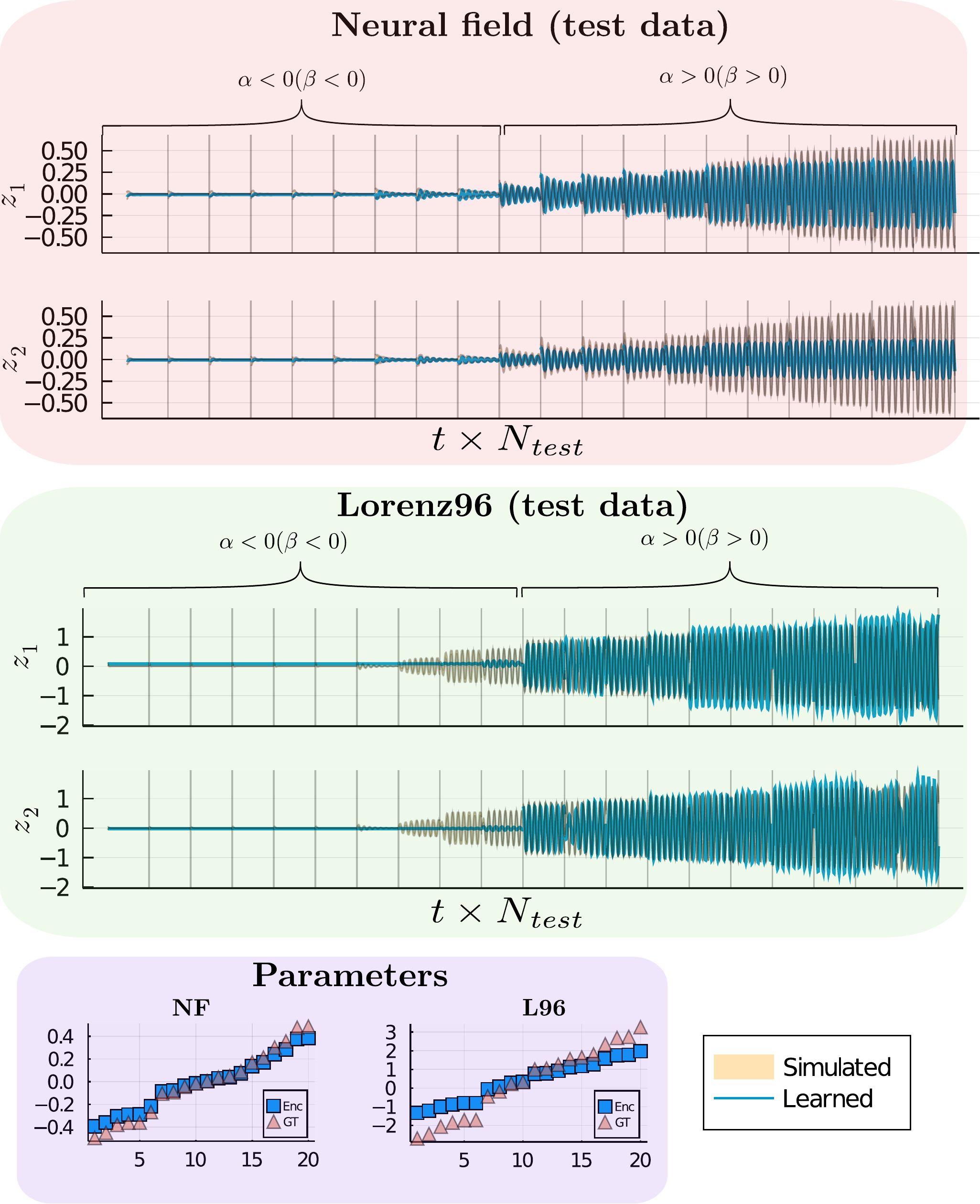}
    \caption{Validation results post-training for the neural field and Lorenz96 examples. The learned normal form coordinates computed via the formula $\bd z = \varphi_1(\bd U)$ is plotted against time $t$ (in blue) for several simulations $N_{\it test}=20$ in the validation set (test data). The different simulations are separated from each other by a vertical gray line. The simulation of the respective normal form is plotted in the background in yellow, which represents an ensemble of 20 trajectories with initial values chosen from a uniform distribution around the first time point $\varphi_1\bd u_0$ and parameter $\beta=\varphi_2\alpha$. For each example the corresponding ground truth parameters ($\alpha$, in blue) and learned parameters ($\beta$, in orange) are also shown. Note that the bifurcation parameter $\alpha_c$ is translated to 0 prior to training. In the Neural field example, the square root effect of the Hopf amplitude fades away for large $\alpha>0$, possibly due to the original parameter $\alpha$ being far from the bifurcation point. }
    \label{fig:suppnfl96}
\end{figure}

\vspace{3mm}
The system is solved with $1000$ initial conditions $(\bd u,\alpha)$ over a temporal domain $[0,100]$ with 250 time points per simulation. The first $50$ points correspond to transients to the periodic breather solution and are removed. The training set thus comprises of $1000$ simulations, giving rise to $2\times 10^5$ training samples. A validation set of $20$ simulations is constructed separately. A batch of 250 simulations is used for each training iteration, giving rise to 2 training iterations per epoch. The system is then trained for 2000 epochs on the training set, or $4000$ training iterations. Training results for both Neural field and Lorenz96 cases are shown in Fig.~\ref{fig:suppnfl96}.
\subsubsection*{Fluid flow past a cylinder (Navier Stokes)}
The final example leverages the model decomposition technique proper orthogonal decomposition (POD) on a high dimensional dataset of fluid flow past a cylinder constructed by solving the Navier Stokes PDE on a 2D domain, to obtain a reduced order dataset on which the normal form autoencoder is trained. The PDE is given by,
\begin{equation}
%    \begin{aligned}
\nabla \cdot \boldsymbol{u} =0, \,\,\,\, 
\partial_{t} \boldsymbol{u}+(\bd u \cdot \nabla)\bd u =-\nabla p+\frac{1}{R e} \Delta \boldsymbol{u},
\label{eq:suppns}
%\end{aligned}
\end{equation}
where $\boldsymbol{u}$ is the two-component flow velocity field in 2D and $\boldsymbol{p}$ is the pressure term. For Reynold's number $Re = Re_c \approx 47$, the fluid flow past a cylinder undergoes a supercritical Hopf bifurcation, where the steady flow for $Re<Re_c$ transitions to unsteady vortex shedding~\cite{Bearman69}. We analyse the one component vorticity field $\bd w$ for the remainder of the work, given by,
\begin{equation}
\bd w = \nabla \times \bd u.
\end{equation}
The training set is formulated in three steps:
\begin{itemize}
    \item Simulate system~(\ref{eq:suppns}) for several initial conditions $(\bd u, Re)$ and generate dataset $\bd U^{(j)}, j=1,2,\dots$ and compute vorticity $\bd W$.
    \item For each simulation, obtain a reduced order dataset $\bd \Lambda$  by projecting the solution $\bd U^{(j)}$ onto finitely many POD modes.
    \item Perform a linear transformation of $\bd \Lambda$ to `mix' the ordered set of harmonics $\bd \Lambda$.
\end{itemize}
\begin{figure}
    \includegraphics[width=0.6\textwidth]{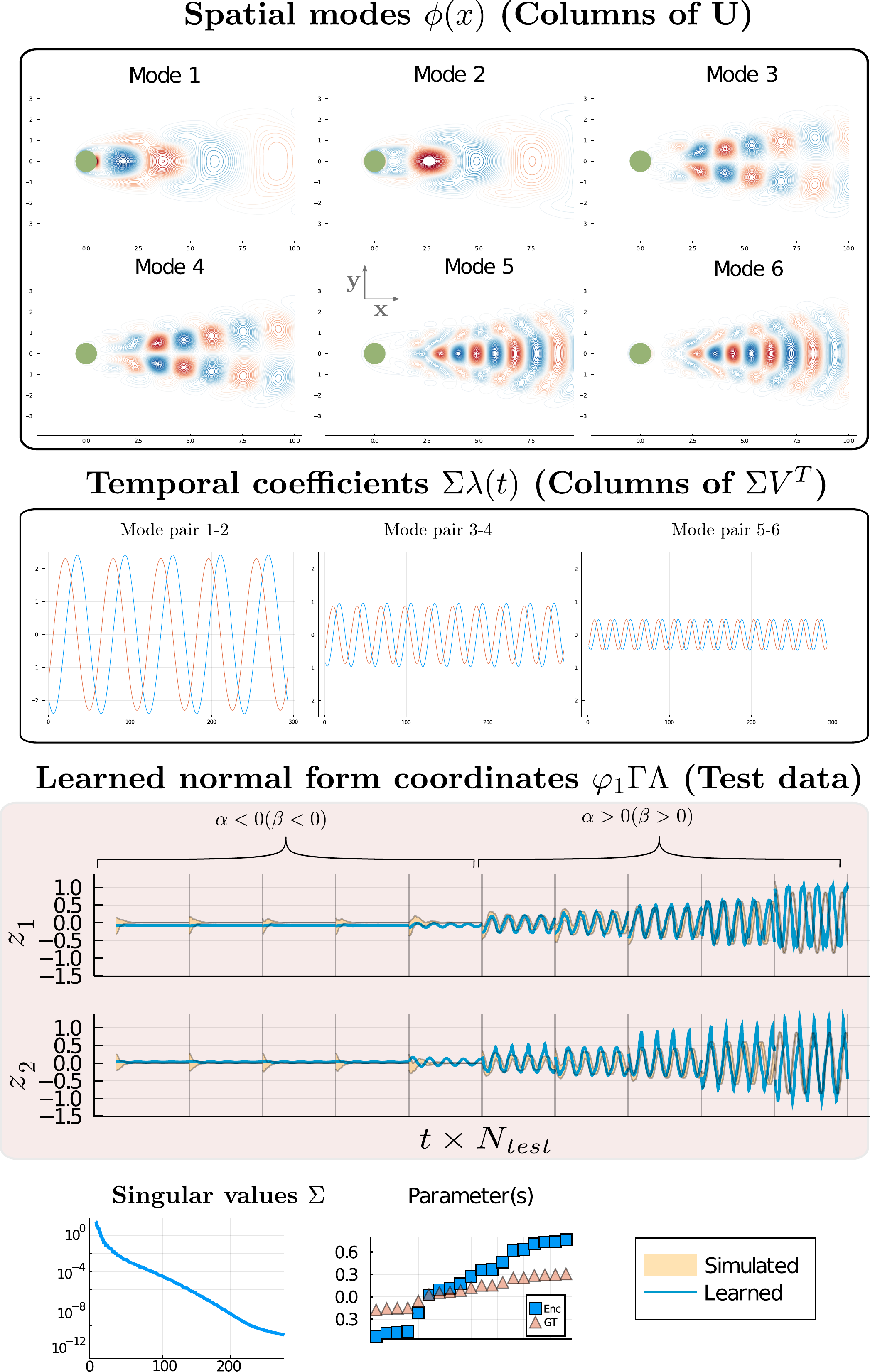}
    \caption{POD calculations and validation results post-training for the fluid flow example (Navier Stokes). In the top two rows, the POD spatial modes and their temporal coefficients are shown, as computed via SVD. Next, the learned normal form coordinates computed via the formula $\bd z = \varphi_1(\bd \Gamma \Lambda)$ is plotted against time $t$ (in blue) for several simulations $N_{\it test}=10$ in the validation set (test data). The different simulations are separated from each other by a vertical gray line. The simulation of the respective normal form is plotted in the background in yellow, which represents an ensemble of 20 trajectories with initial values chosen from a uniform distribution around the first time point $\varphi_1\bd u_0$ and parameter $\beta=\varphi_2\alpha$. The corresponding ground truth parameters ($\alpha$, in blue) and learned parameters ($\beta$, in orange) are also shown.}
    \label{fig:suppfluid}   
\end{figure}
\textbf{Simulation.} The Navier-Stokes PDE is solved for 250 initial values $(\bd u,Re)$ centered around the critical point $Re_c \approx 44.6$ on the spatial domain $x \times y = [-2,10]\times [-3,3]$ with $487 \times 250$ spatial grid points. The choice for $\Delta x, \Delta y$ and $\Delta t$ is made by using the cell Reynold's number of $1.3$ for $Re=80$, and calculating the appropriate CFL condition after setting $\Delta t$. The temporal domain is $[0,77]$ comprising of 6180 time steps per simulation. The cylinder is represented as a circle centered at $(x,y) = (-1,1)$ with diameter $1$. Equation~(\ref{eq:suppns}) is solved in voriticity form using the immersed boundary projection method \cite{Taira2007} which is implemented in the Julia package \texttt{ViscousFlow.jl}. This procedure gives us the dataset $\bd W^{(j)}$.

\vspace{3mm}
\textbf{Projection onto POD modes.} Proper orthogonal decomposition is a model decomposition technique that constructs a set of spatial basis functions in descending energy, from which a reduced order dataset can be created by projecting only onto a few high-energy modes \cite{Rowley2004}. Thus, a solution $\bd w (\bd x,t)$ is written as a Galerkin projection onto POD modes $\bd \phi(\bd x)$ and their evolving temporal coefficients $\bd \lambda(t)$ as follows,
\begin{equation}
    \bd w(\bd x,t) = \bar{\bd w} + \sum_k \sigma_k\bd \phi(\bd x)_k \lambda_k(t),
\end{equation}
where $\bar{\bd w}$ represents the mean flow $\mathbb{E}_t\bd W$. The coefficients $\sigma_k$ form a descending sequence of singular values. In practice, the spatial modes are computed using the `method of snapshots' \cite{Sirovich87} implemented by singular value decomposition (SVD). Thus, for a simulation $\bd W^{(j)}$, SVD yields unitary matrices $U, V$ and a diagonal matrix $\Sigma$ such that 
\begin{equation}
    \bd{W}^{(j)} = U\Sigma V^T.
\end{equation}
The columns of matrix $U$ represent the spatial modes, for which the columns of $V$ give the temporally evolving coefficients. The matrix $\Sigma$ is composed of $t_f$ singular values in descending sequence, which is significantly smaller than the state space dimension (121750). Choosing the first $m$ singular values generates an approximate reduced order model,
\begin{equation}
\bd{W}^{(j)} \approx U_m\Sigma_mV_m^T,
\end{equation}
where $U_m$ and $V_m$ are truncated matrices composed of the first $m$ columns of $U$ and $V$, respectively. We work with $m=4$, which gives a reduced-order model $V$ of dimension 4. The SVD is performed in four steps:
\begin{itemize}
    \item First the transients to the vortex shedding solution or the planar flow solution are trimmed off the first 3250 points.
    \item Next, the mean flow of the trimmed solution is computed and subtracted from the simulation.
    \item Finally, POD is permed via the method of snapshots, but on a dilated temporal scale, using $\Delta t \mapsto 10\Delta t$, as done in \cite{kutz2016dynamic}. This yields matrices $U,V$ and $\Sigma$, where $V$ has dimension $293 \times 293$. 
    \item The truncation above is performed on the matrices obtained via SVD, to obtain the dynamical system $V$ with dimension $4\times 293$.
\end{itemize}
The reduced order dynamical system $\bd \Lambda^{(j)}$ is then given by 
\begin{equation}
    \bd \Lambda^{(j)} = \Sigma_m^{(j)}(V^{(j)})^T,
\end{equation}
where the superscript $(j)$ indicates index of a specific simulation.

\vspace{3mm}
\textbf{Linear transformation of $\bd \Lambda$}. The resulting dynamical system $\Lambda$ is composed of rows of pairwise harmonics that increase in frequency and decrease in amplitude for an increasing number of rows. Following suggestions from \cite{Noack2003jfm} on constraining Galerkin models to the Stuart-Landau expression (Hopf normal form), we introduce a linear transformation $\Gamma$ of $\Lambda$, before training. This preserves the original frequency of the periodic orbit, but has the added disadvantage of introducing multiple timescale dynamics in the periodic orbit, which can be hard to remove in the latent space dynamics. $\Gamma$ is chosen randomly, but in such a way that its condition number is close to 1. In other words, we choose $\Gamma$ to be unitary. This choice has the advantage of obtaining better reconstruction when projecting back into the original 2D space, as the higher-order harmonics in $\Lambda$ are preserved. This is done by first generating a random matrix $\tilde \Gamma$, and obtaining $\Gamma$ via SVD,
\begin{align*}
    \tilde \Gamma &= U \Sigma V^T, \\
    \Gamma &= U V^T.
\end{align*}
In the results we show in the manuscript, this matrix $\Gamma$ is given by,
\begin{equation}
    \Gamma = \begin{bmatrix}
        0.154739 & -0.523688 & 0.675546 & 0.495243 \\
        0.87244 & 0.298319 & 0.249166 & -0.29685 \\
        -0.292797 & 0.785392 & 0.450626 & 0.30719 \\
        0.359894 & 0.141123 & -0.527721 & 0.756353
    \end{bmatrix}.
\end{equation}
\textbf{Reconstruction post training.} The dynamical system $\Gamma \Lambda^{(j)}$ is used for training the normal form autoencoder. The spatial modes $U_m^{(j)}$ and the mean solution $\bar{\bd w}^{(j)}$ are stored offline, and are used to reconstruct the full simulation $\hat{\bd W}^{(j)}$ via the relation
\begin{equation}
    \hat{\bd W}^{(j)} = \bar{\bd w} + U^{(j)}\cdot \Gamma^T\hat{\bd \Lambda}^{(j)},
\end{equation}
where $\hat{\bd \Lambda}^{(j)}$ is the projection of the latent space onto the larger dimensional space via the state decoder $\psi_1$,
\begin{equation}
    \hat{\bd \Lambda}^{(j)} = \psi_1(\varphi_1\bd \Gamma\Lambda^{(j)}). 
\end{equation}
POD modes and their projections onto the normal form coordinates are shown in Fig.~\ref{fig:suppfluid}.

%\newpage
%\bibliographystyle{unsrt}
%\bibliography{ref.bib}
%\end{document}
%

%

%\clearpage
\small
\bibliographystyle{unsrt}
\bibliography{ref.bib}

\newpage

\clearpage
%\includepdf[pages=-]{supp.pdf}

\end{document}